\documentclass[a4paper,twoside]{article}

\usepackage{epsfig}
\usepackage{graphicx}
\usepackage{wrapfig}
\usepackage{xcolor}
\usepackage{color,colortbl}
\usepackage{subcaption}
\usepackage{calc}
\usepackage{amssymb}
\usepackage{amstext}
\usepackage{amsmath}
\usepackage{amsthm}
\usepackage{multicol}
\usepackage{pslatex}
\usepackage{apalike}
\usepackage{microtype}
\usepackage{bm}
\usepackage{flushend}
\usepackage{adjustbox}
\usepackage{SCITEPRESS}     

\definecolor{Gray}{gray}{0.9}
\begin{document}

\title{Deep Learning Based Model Identification System Exploits the Modular Structure of a Bio-Inspired Posture Control Model for Humans and Humanoids}

\author{\authorname{Vittorio Lippi\sup{1}\orcidAuthor{0000-0001-5520-8974}}
\affiliation{\sup{1}Neurological University Clinic, University of Freiburg, Freiburg im Breisgau, Germany}
\email{vittorio.lippi@uniklinik-freiburg.de} 
}


\abstract{This work presents a system identification procedure based on Convolutional Neural Networks (CNN) for human posture control using the DEC (\textit{Disturbance Estimation and Compensation}) parametric model. The modular structure of the proposed control model inspired the design of a modular identification procedure, in the sense that the same neural network is used to identify the parameters of the modules controlling different degrees of freedom. In this way the presented examples of body sway induced by external stimuli provide several training samples at once.}

\onecolumn \maketitle \normalsize \setcounter{footnote}{0} \vfill

\section{\uppercase{Introduction}}
\label{sec:introduction}
The application of \textit{convolutional neural networks} (CNN) in human movement analysis has produced promising results in recent experiments \cite{batchuluun2018body,karatzoglou2018convolutional,icinco19,icinco19Abdu}, and in general deep learning is starting to be applied to system identification \cite{de2016randomized,andersson2019deep,miriyala2020deep,de2015nonlinear,ljung2020deep}. This work aims to apply CNN to the identification of human posture control models, that are used in several studies for the analysis of human data and the control of humanoids, e.g. \cite{vanderKooij2007,van2005comparison,van2006disentangling,goodworth2018identifying,Mergner2010,engelhart2014impaired,pasma2014impaired,jeka2010dynamics,boonstra2014balance}. Most of the studies on posture control use linear models like the \textit{independent channel} (IC) model \cite{peterka2002sensorimotor}, and make the assumption of linear and time-invariant behavior \cite{engelhart2016comparison}. In this work a nonlinear model will be considered. This makes the use of deep learning more interesting, as the identification of nonlinear systems is more complex and in general it is performed using computationally expensive iterative algorithms as in \cite{asslander2015visual}. In a previous work, CNNs proved to be suitable to identify the parameters of a human posture control model \cite{icinco20}. Specifically, the CNN presented in \cite{icinco20} was applied to a single inverted pendulum (SIP) model, but it proved to work also in identifying the parameters of a double inverted pendulum system. In this work a triple inverted pendulum (TIP) model of posture control will be used to show how the identification process can be extended to an arbitrary number of degrees of freedom.  The model used here, the \textit{Disturbance Estimation and Compensation} (DEC) has a modular structure \cite{lippi2017human}, i.e. each degree of freedom is controlled by a module and all the modules share the same structure. Exploiting this characteristic, one CNN is used to identify the parameters of the three modules controlling the three degrees of freedom. In this way each simulation provides three training samples at once.  
	
\section{\uppercase{Methods}}
\subsection{Posture Control Scenario}
The scenario considered here models a human (or humanoid) balancing on a tilting support surface as a TIP. The three degrees of freedom considered in the sagittal plane are the ankles, the knees and the hips (Fig. 1A).  The support surface tilt $\alpha_{FS}$ represents the input of the system and it is the same for all the simulations. The profile of the tilt of the support surface is the \textit{pseudo-random ternary sequence}, PRTS, shown in Fig. 1B. Such stimulus is used in human experiments because it is not predictable for the subject \cite{peterka2002sensorimotor}. It is composed by a sequence of velocity steps suitable to excite the dynamics of the system over several frequencies. The output of the system is the sway of the body segments: shank, thigh (leg) and trunk addressed as $\alpha_{SS}$, $\alpha_{LS}$, and $\alpha_{TS}$ respectively.

\subsection{Human and Humanoid Posture Control: The DEC Model}
\begin{figure}[t]
	\centering

		\includegraphics[width=1.00\columnwidth]{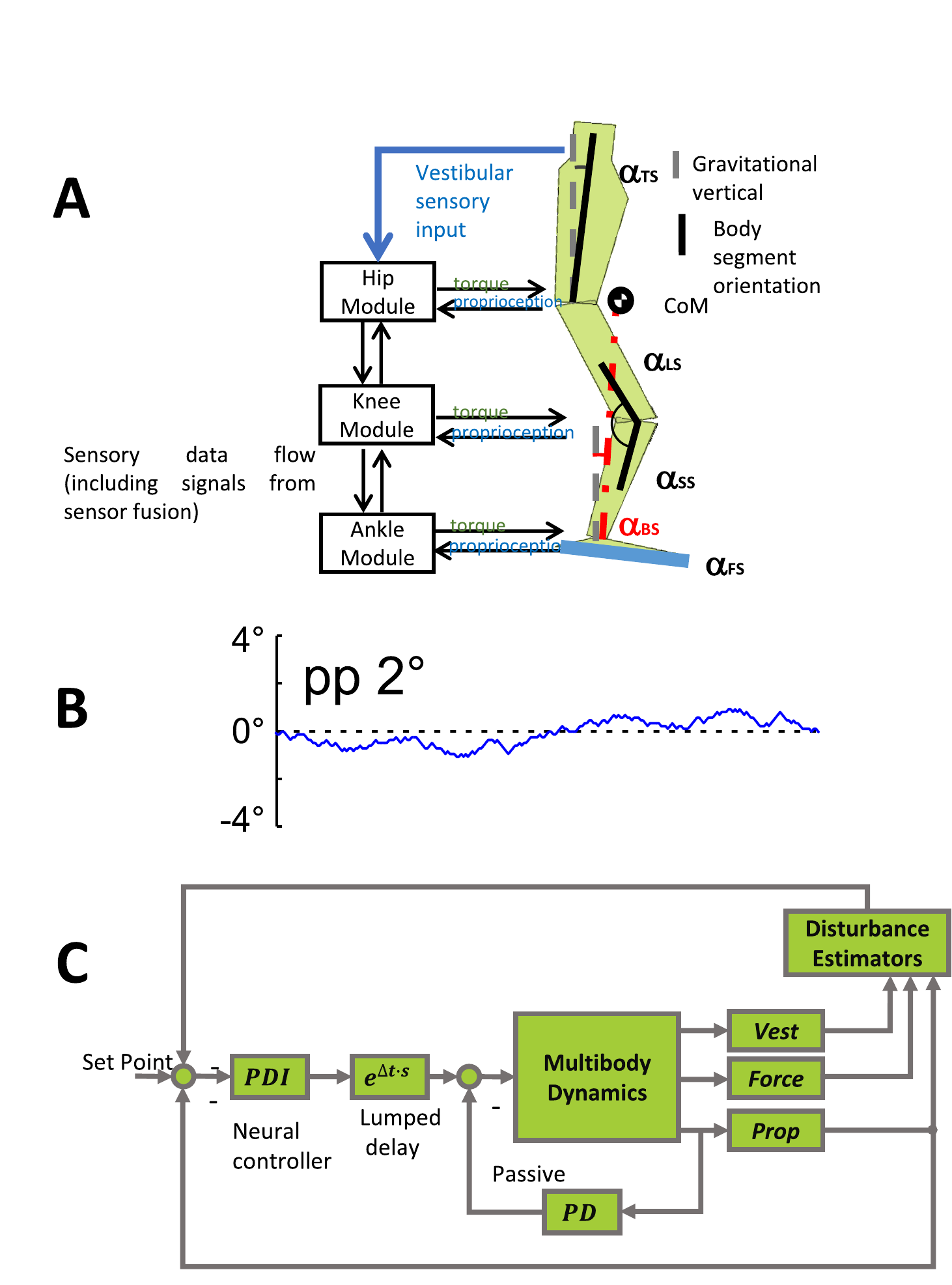} \\ 
		
	\caption{Balance scenario and posture control system. (A) The triple inverted pendulum model used to simulate human posture control together with the control modules. The $\alpha_{BS}$ represent the sway of the body COM around the ankle joint. (B) The \textit{Pseudo-Random Ternary Signal}, PRTS, the time profile for support surface tilt. (C) Schema of a control module. The disturbances are compensated feeding them in the form of an \textit{angle equivalent} as input to the servo controller (PID).} 
	\label{DEC}
\end{figure}

The DEC is a model of the human postural control mechanisms \cite{Mergner2010,lippi2017human}. It has a modular structure that can be applied to multiple DoF robots \cite{lippi2017human,zebenay2015human,ott2016good,lippi2018prediction,Hettich2013,hettich2015human}. Here it is implemented specifically on the TIP model, using three control modules. A block diagram of the DEC control is shown in Fig.\ref{DEC}. A complete description of the model is beyond the scope of this paper and can be found in \cite{lippi2017human}. In the following a general description of the model is provided with particular emphasis on the components that have an impact on the definition of the machine learning problem.
A control module based on the DEC concept is implemented as servo loop, here implemented as a PID controller (the neural controller in Fig. \ref{DEC}C). The controlled variable can consist in the COM sway of all the segments above the joint controlled by the module with respect to the gravitational vertical passing through the controlled joint (e.g. $\alpha_{BS}$ in Fig \ref{DEC} for the ankle joint). The sensory channels shown in Fig.\ref{DEC} (C) as \textit{Vest}ibular, \textit{Prop}rioceptive, and \textit{Force} are used for disturbance estimation. The disturbance estimates are fed into the servo so that the joint torque on-line compensates for the disturbances. The \textit{lumped delay} in Fig. \ref{DEC} (C) represents  all the delay effects that are distributed in humans, and humanoids \cite{antritter2014stability,Hettich2014}. The model used in this work considers gravity and support surface tilt as disturbances. The gravity torque to be compensated is assumed to be proportional to the sway of the COM of groups of segments above the controlled joints. Ideally the gain required to compensate gravity is $mgh$, where $m$ and $h$ are the total mass and the COM height of the controlled segments respectively, and $g$ is the gravity acceleration. The integral component of the PID controller is not applied to the compensation as in \cite{ott2016good}. The support surface tilt estimator includes a non-linearity, introduced to represent the behavior observed in humans \cite{T.Mergner2009,Mergner2003}, defined as:
\begin{equation}
	\alpha_{FS}=\int_0^t f_{\theta} \left( \frac{d}{dt} \alpha_{BS}^{vest} + \frac{d}{dt} \alpha_{BF}^{prop} \right)
	\label{fs}
\end{equation}
where $\alpha_{BF}^{prop}$ is the ankle joint angle signal from proprioception. $BF$ stands for \textit{Body-to-Foot}. The function $f_{\theta}$ is a dead-band threshold defined as
\begin{equation}
	f_{\theta}(\alpha) = \left\{ 	
	\begin{array}{llc}
   		\alpha + \theta & if & \alpha< -\theta \\
			0 & if & -\theta< \alpha< \theta \\
			\alpha - \theta & if & \alpha> \theta \\
 	\end{array}
	\right.
\end{equation}
The estimated $\alpha_{FS}$ is then \textit{up-channeled} through the control modules and used to control body position. In this work the threshold is set as $\theta = 0.0003 rad/s$.
The disturbance compensation and  the total torque commanded by the servo controller for the ankle joint is:
\begin{equation}
	\tau(s)=-e^{-s\Delta } \left(K_p+s K_d+K_i\frac{1}{s}\right)\left(\epsilon \right) + (K_p+s K_d) T_g
	\end{equation}
where $K_p$ and $K_d$ is the derivative coefficient for the PD controller, and  $\Delta$ is the lump delay. Notice that the derivative component is also acting  on gravity $T_g$, representing a sort of anticipation of the disturbance. There is also a passive torque acting on the joints represented as a spring-damper model:
\begin{equation}
	\tau_{passive}=-(K_p^{pass}+s K_d^{pass})\left(\alpha_{BF}^{prop}\right)
	\end{equation}
The coefficients $K_p^{pass}$ and $K_d^{pass}$ are kept fixed for all the simulations. The standard parameters are shown in Table \ref{tab:parameters}. In general the gains are proportional to the $mgh$ for the groups of segments above the controlled joint, this will be used in the next section to normalize the parameters. The anthropometric of the TIP model is taken from previous works \cite{lippi2019distributed,lippi2020challenge}.

\subsection{The Training Set}
\label{data}
\paragraph{Control parameters and target set.} The training set has been generated simulating the posture control scenario with different parameters, these parameters represent the target for the neural network and the sway of the segments represents the input. In order to exploit the modular structure of the DEC control the CNN has been designed to identify the parameters of a specific module. In this way each simulation provides three training samples associated with the three control modules. As explained in the previous section, the gains of the controllers are proportional to the $mgh$ of the segments above the controlled joint, and the lumped delay is larger for the ankle module and smaller for the hip module. In order to use such parameters for the training of the same CNN their value is expressed as deviation from the default value and normalized dividing it by the default value, e.g. $\tilde{K_p}=(K_p^{sample}-K_p)/K_p$ in Table \ref{tab:parameters}. Besides the PID gains the target vector includes the lumped delay and a variable $C$ that is set to $-1$ when the controlled variable is the joint angle and to $+1$ when the controlled variable is the COM. This leads to a sample with the following form:
\begin{equation}
	\tilde{T}=\left[\tilde{K_p}\ \ \tilde{K_i}\ \ \tilde{K_d}\ \ \tilde{\Delta} \ \ C  \right]
\end{equation}
 where the $\tilde{ }$ indicates that the values are normalized. 

\begin{figure}[htbp]
	\centering
		\includegraphics[width=1.00\columnwidth]{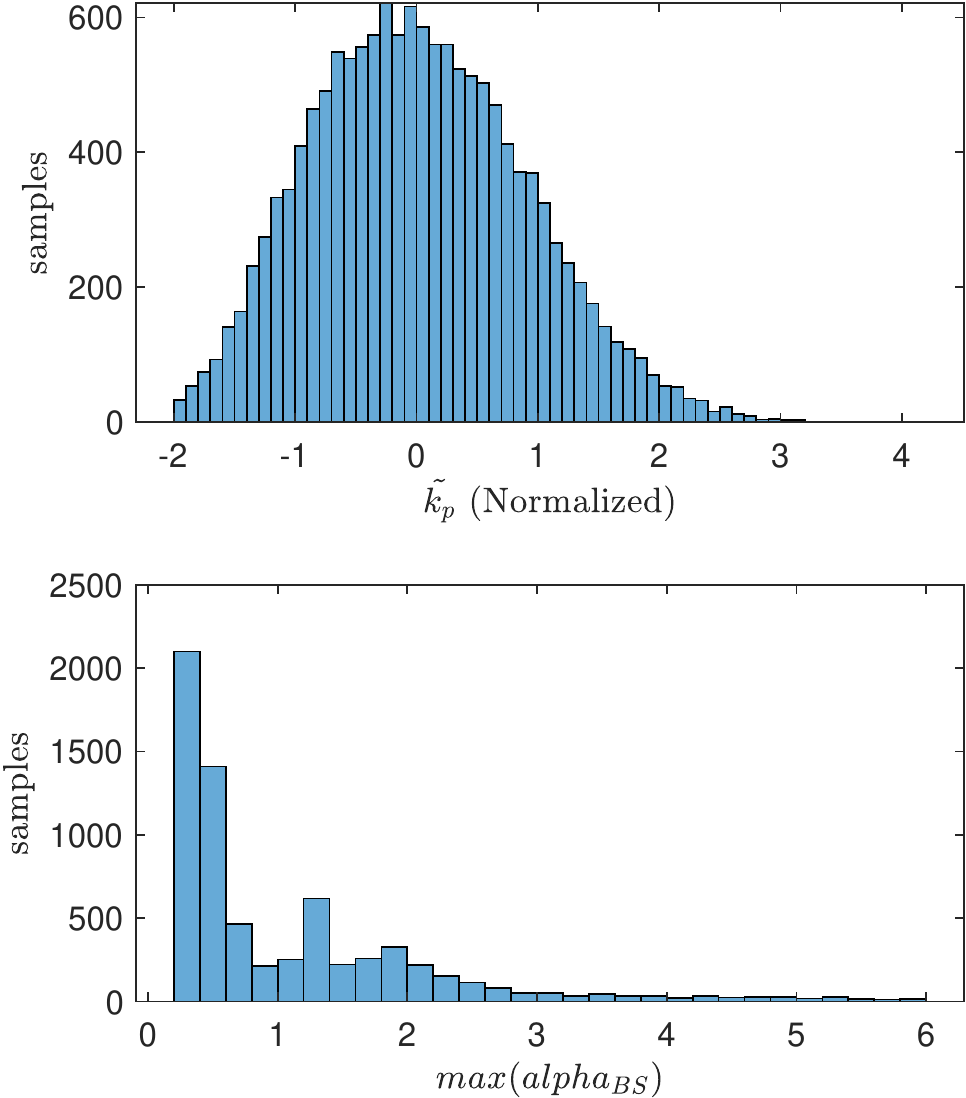}
	\caption{Output of the performed simulations. Above the distribution of the normalized $\tilde{K_p}$ parameter produced during the stable simulations. Below the distribution of the body sway amplitudes. The last column of the histogram on the right includes the simulations that were aborted because the body sway reached the threshold of $6^{\circ}$.}
	\label{fig:histograms-cropped}
\end{figure}

The training samples are generated with random parameters from normal distributions $X \sim \mathcal{N}(0,\,0.5)$ that is summed to the normalized parameters. The variable $C$ is sampled with equal probability ($0.5$) between the two cases. In order to avoid negative values the absolute value of the obtained parameters is used "`warping"' the normal distribution on positive values. A set of parameters is used as a sample only if the behavior it produces is stable: simulations with body sway larger than $6^{\circ}$ are not considered realistic balancing scenarios and are discarded. Overall the obtained data consist of $20730$ samples, $14000$ used as training set, $3000$ as validation set and $3730$ as test set.
The three sets are normalized subtracting the average divided by the variance of the training set. The resulting distribution of the normalized $\tilde{K_p}$ is shown in Fig. \ref{fig:histograms-cropped} together with the distribution of maximum body sway amplitudes.

\begin{table*}[t!]
\center
\renewcommand{\arraystretch}{2}
\begin{tabular}{|l||l|l|l|l|l|}
\hline
Parameter                        & Symbol         & Ankle      & Knee       & Hip    & unit                            \\ \hline \hline
\rowcolor{Gray}
Active proportional gain         & $K_p$          & 465.98 & 245.25 & 73.575 &  $\frac{N \cdot m}{rad}$         \\
Active derivative gain           & $K_d$          & 116.49 & 18.394  & 18.394 &  $\frac{N \cdot m \cdot s}{rad}$ \\
\rowcolor{Gray}
Active integrator gain           & $K_i$          & 11.649 & 6.1312  & 1.8394 &  $\frac{N \cdot m }{rad \cdot s}$ \\
Passive stiffness                & $K_{p_{pass}}$   & 232.5000  & 61.2500  & 36.5000 &  $\frac{N \cdot m}{rad}$         \\
\rowcolor{Gray}
Passive damping                  & $K_{d_{pass}}$   & 145.000  & 11.2500  & 11.2500 &  $\frac{N \cdot m \cdot s}{rad}$ \\
Foot rotation velocity threshold & $\theta_{vfs}$ & 0.03       &     &    &  $rad / s$                       \\ 
\rowcolor{Gray}
Lumped delay                     & $\Delta$       & 0.10        & 0.07    & 0.1210   &  $s$                             \\ \hline
\end{tabular}
\caption{Default parameters}
\label{tab:parameters}
\end{table*}

\begin{figure}
	\centering
		\includegraphics[width=1.00\columnwidth]{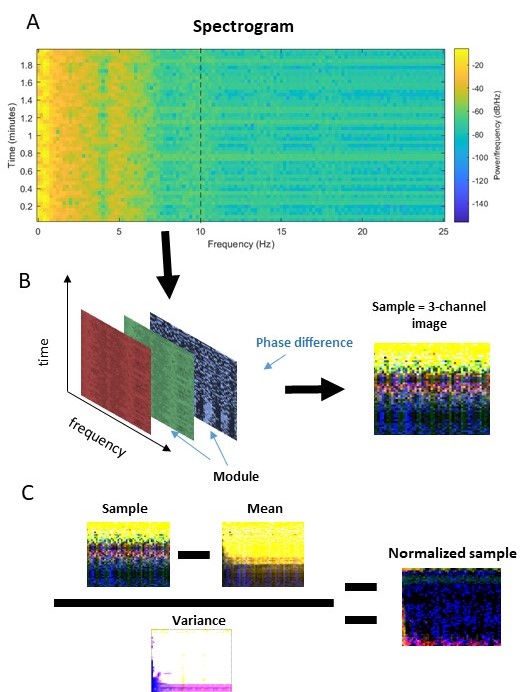}
	\caption{Design of the input features for the CNN. In A the spectrogram of a signal is given, the dashed line shows where it is cut. In B it is shown how the spectrograms associated with the sway of two body segments are combined to form a three channel $51 \times 51$ pixel image. The image  C shows the normalization of the samples using the mean and the variance of the training set, computed element-wise (pixel by pixel).}
	\label{fig:IF3c}
\end{figure}

\paragraph{The CNN input.} The \textit{Input} is  a representation of the system output, i.e the sway of the segments. The simulation was performed with a fixed integration step of $2 ms$ and sampled at $50 \ Hz$, producing $6051$ samples with a resolution of $20 \; ms$. In order to adapt the signal to the CNN the input was transformed into an image. The data considered were the sway with respect to the vertical of the two segments under and above the controlled joint, e.g. for the ankle module they are $\alpha_{FS}$ and $\alpha_{LS}$. The Spectrogram of the two signals is computed with short-time Fourier transform over windows of 250 samples overlapping for 135 computing 250 frequency points. This results in a $51 \times 250$ matrix of complex values. The first 51 columns of the matrix are here considered, corresponding to a bandwidth of approximately $ 10 \ Hz$. Due to the low-pass dynamics of the body sway, the power spectrum of the signal at higher frequency is very poor and hence not giving important informations (i.e. almost black images independently of the parameters) . The resulting $51 \times 51$ matrix is used to define the input image as follows: the module and the phase of the matrices associated to the two segments sway are computed. The first channel of the image is the module of the matrix describing the sway of the segment above, the second is the matrix describing the segment below and the third channel is the difference between the phases. The process is summarized in Fig \ref{fig:IF3c}.
Also the input images are normalized subtracting the average and dividing by the variance of the training set through element-wise operation (i.e. pixel by pixel, channel by channel). 

\subsection{CNN Architecture}

\begin{figure*}[t!]
	\centering
		\includegraphics[width=1.00\textwidth]{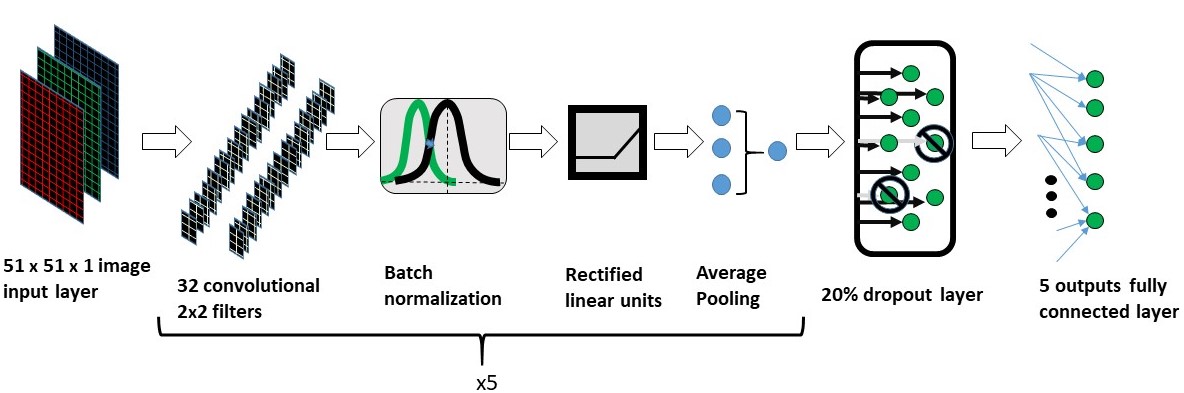}
	\caption{Neural Network architecture. The group of layers addressed with \textit{x 5} is repeated 5 times.}
	\label{fig:Neuralarchitecture}
\end{figure*}

The neural network architecture is schematized in Fig \ref{fig:Neuralarchitecture}.  The network is implemented with {Matlab\texttrademark} Deep Learning Toolbox\texttrademark.  The network has been trained using \textit{stochastic gradient descent with momentum} as policy. The training was set to a limit of 200 epochs. The loss function is the \textit{Mean Squared Error} MSE as expected with a regression task, although the target vector includes a categorical feature, the controlled variable, that implies a classification problem. The categorical variable produced by the CNN is interpreted considering the sign (positive = COM sway, negative = joint angle). The performance in regression and classification is discussed separately in the Results section. 

\section{Results}
\subsection{Training and Test}
\begin{figure}[htbp]
	\centering
		\includegraphics[width=1.00\columnwidth]{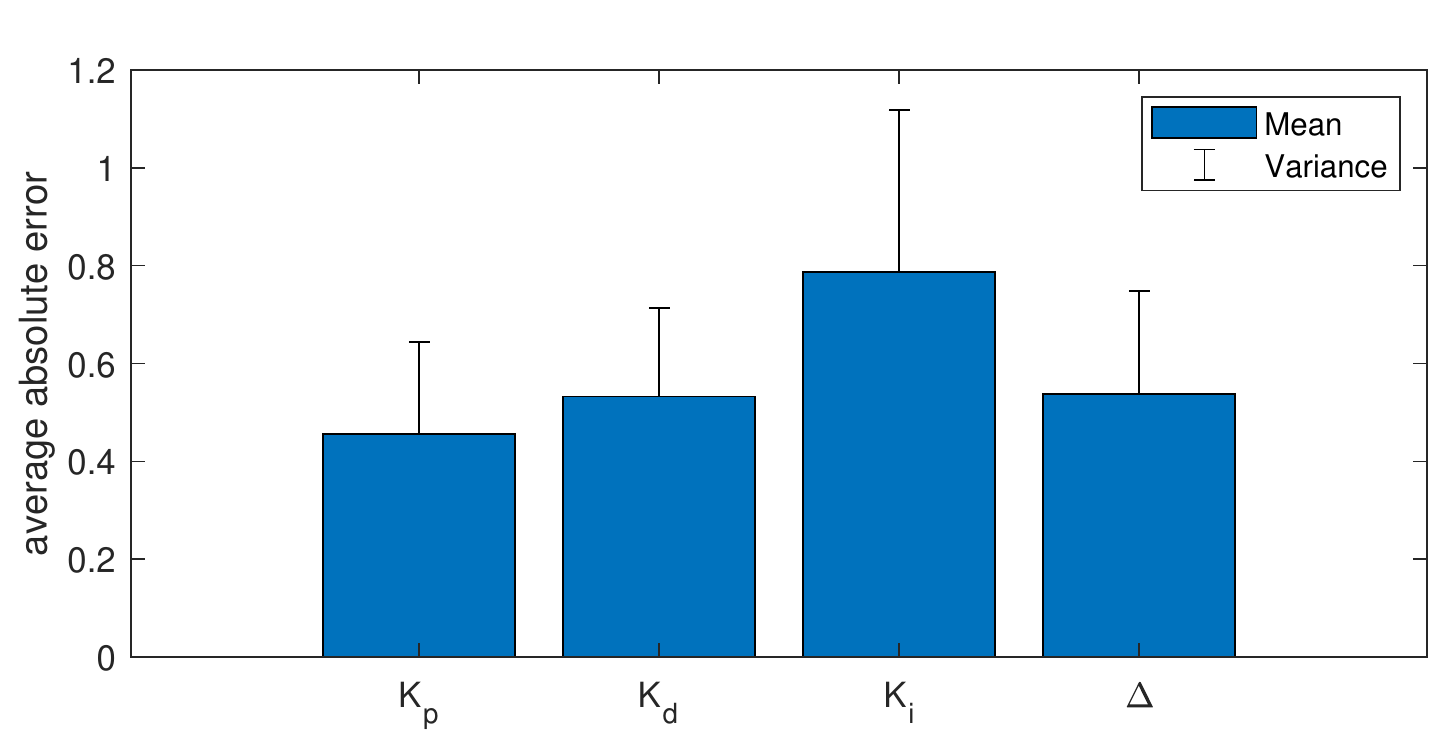}
	\caption{Average and variance of the error on the parameters.}
	\label{fig:ErrorsCrop}
\end{figure}

\begin{figure}[htbp]
	\centering
		\includegraphics[width=1.00\columnwidth]{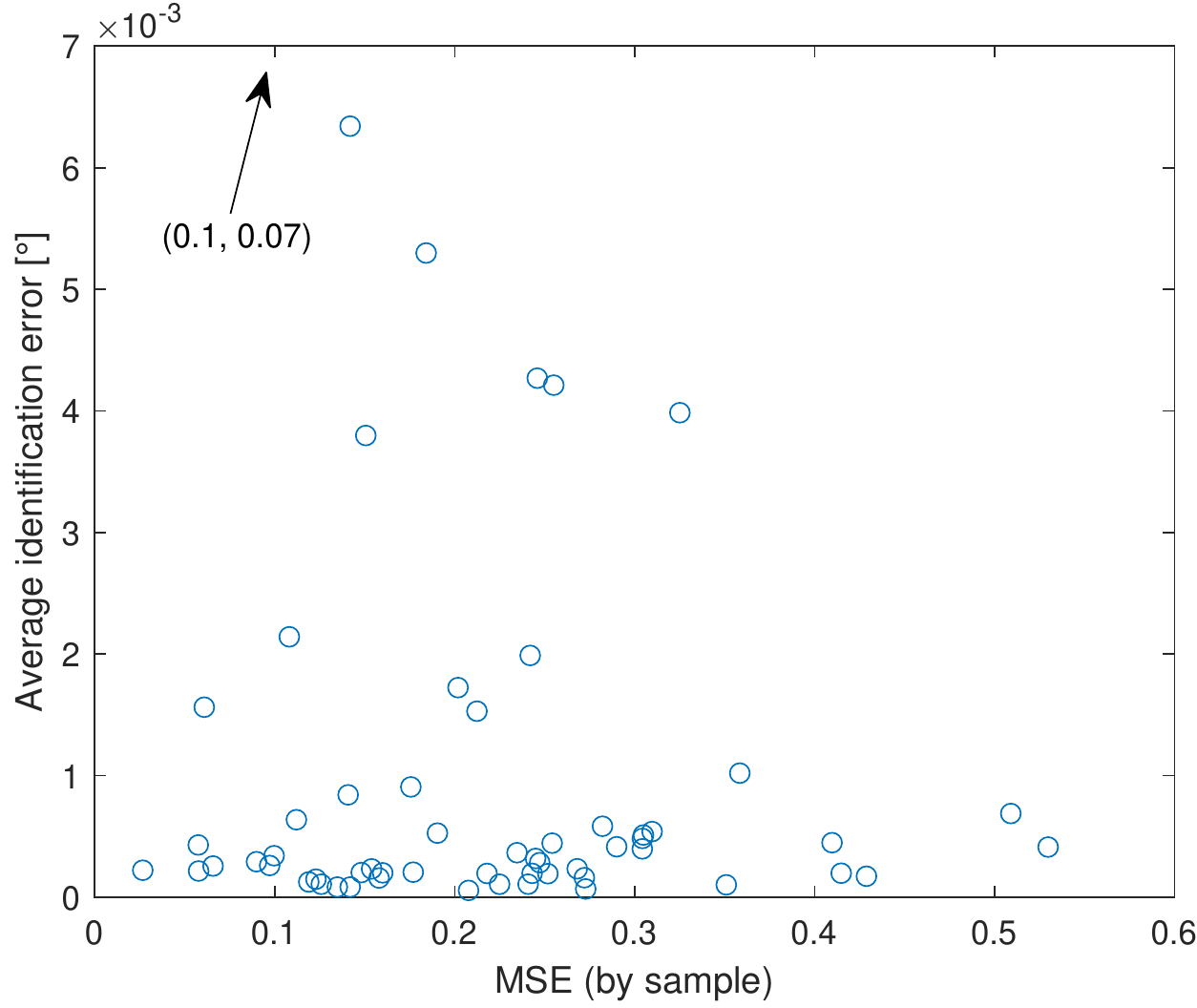}
	\caption{Identification Error for a set of 60 training samples. The identification error is plotted versus the MSE error on the parameters (target output). The arrow addresses a point that is out of the scale of the axes (i.e MSE = 0.1, error=0.07). The correlation between the two measures of errors is $-0.1265$, which means in general larger MSE on the parameters are not associated to the larger identification errors. This may suggest a certain degree of redundancy in the model, allowing the system to reproduce a similar output with different sets of parameters.}
	\label{fig:idERROR}
\end{figure}

The results of the training set are reported in Table \ref{tab:TrainingResults}. Notice that there is a total RMSE on the training set, including the error on the categorical variable, and a fit RMSE computed on the continuous variables. The average absolute error and absolute variance is shown in Fig. \ref{fig:ErrorsCrop}. The error is larger on $K_i$. The identification error in this context can be defined as the norm of the difference between body sway obtained with the target parameters ($\bm{\alpha}_{BS}$) and the one associated with the identified parameters ($\bm{\tilde{\alpha}}_{BS}$) divided by the number of samples $N=6051$,
\begin{equation}
	E_{id}=\frac{\left\|\bm{\alpha}_{BS}-\bm{\tilde{\alpha}}_{BS}\right\|}{N}
\end{equation} ,
where the bold text represents the fact that $\mathbf{\alpha}_{BS}$ is a vector of samples. The MSE on the single sample is computed considering the target error on the three control modules used in the simulation. The prediction error $\epsilon_{p}$ on the 15 normalized parameters is computer for each sample, leading to an MSE of
\begin{equation}
	MSE=\sqrt{\epsilon_{p}^T \epsilon_{p}}/15
\end{equation} .
The identification error plotted versus the MSE is shown in Fig. \ref{fig:idERROR}. On average the identification error on the training set is $0.0024^{\circ}$

\begin{table}[h]
	\centering
	\begin{adjustbox}{max width=1\columnwidth,center}
		\begin{tabular}{|c||c|c|c|}
			\hline
			\textbf{Dataset} & \textbf{total RMSE} & \textbf{fit RMSE} & \textbf{Accuracy} \\
			\hline
			\hline
			\textbf{Training} & 1.3664 & 1.3296  & 85.79\% \\
			\hline
			\textbf{Validation} & 1.5186 & 1.4810 & 83.90\% \\
			\hline
			\textbf{Test }& 1.4846 & 1.4486 & 84.72\% \\
			\hline
		\end{tabular}
		\end{adjustbox}
	\caption{Training Results}
	\label{tab:TrainingResults}
\end{table}
\normalsize

\subsection{Identification of Human Posture Control Parameters}
\begin{figure}[htbp]
	\centering
		\includegraphics[width=1.00\columnwidth]{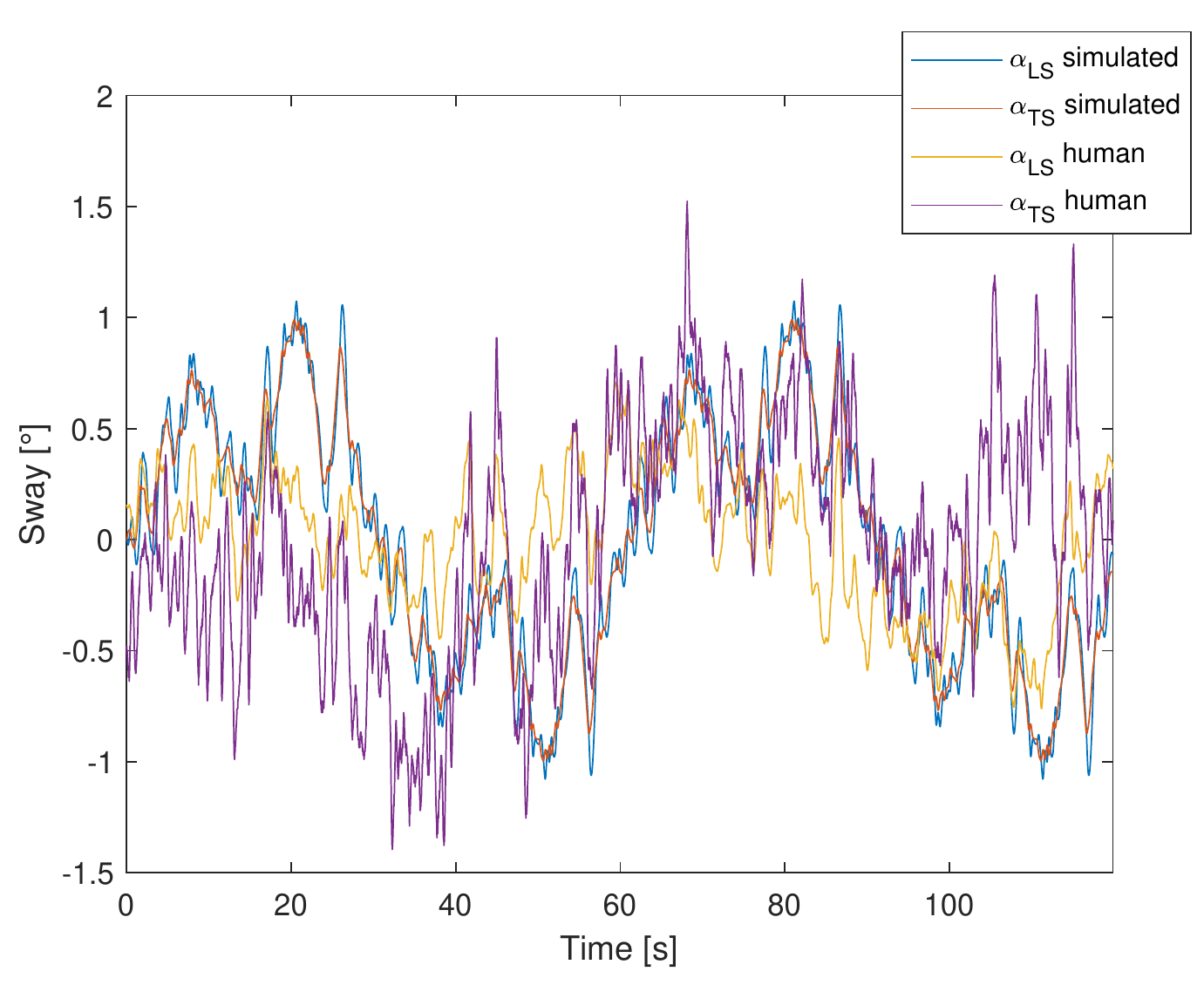}
	\caption{Human data compared with simulated responses.}
	\label{fig:Human}
\end{figure}

The neural network is used to identify the control parameters from a human experiment performed with the same scenario simulated in the training set. The human experiment had the user stand with straight legs and hence the simulation was applied just to the ankle and the hip module.
The identified parameters are reported in table \ref{human} and the simulated sways of the segments are shown in Fig. \ref{fig:Human},  the result shows a good similarity between the simulation and the original data. For both the control modules, i.e. hip and ankle, the network proposed the sway in space of the segments as controlled variable, in according to previous studies on support surface tilt \cite{Hettich2014}.
\small
\begin{table}[ht]
	\centering
\scriptsize
\begin{adjustbox}{max width=1\columnwidth,center}
		\begin{tabular}{|c||c|c|c|c|c|}
			\hline
			\textbf{Module} & \textbf{$K_p$} & \textbf{$K_i$} & \textbf{$K_d$} & \textbf{$\Delta$} & \textbf{Var} \\
			\hline
			\hline
			\textbf{Ankle} & 421.8574  & 74.6664  & 12.0254  &  0.0685 & Sway Angle\\
			\hline
			\textbf{Hip} &74.5148 &  8.2559   & 1.8854 &   0.0219 & Sway Angle\\
			\hline

		\end{tabular}
		\end{adjustbox}
	\caption{Parameters identified for an example from a human experiment.}
	\label{human}
\end{table}
\normalsize

\subsection{Identification with a \textit{Monolithic} model}
In order to evaluate the impact of the modular approach the experiment is repeated with a network that considers the three DOF together. The target vector incorporates the parameters for the 3 control modules, i.e. 15 values; the input is encoded as a 5 channel picture with with spectrograms as described in \S \ref{data}. Support surface tilt $\alpha_{FS}$ is not taken in account as it is the same for all the samples. The channels represent the modulus of the spectrogram for  $\alpha_{LS}$,  $\alpha_{MS}$, and  $\alpha_{TS}$, and the phase difference between $\alpha_{LS}$ and $\alpha_{MS}$ as well as between $\alpha_{MS}$ and $\alpha_{TS}$. The dataset, rearranged in this way accounted for $2264$ samples, i.e. one third of the original set. The dataset was split into a training set and a validation set of $1000$ samples and a test set of $264$ samples.The network had the same structure of the one used in the modular case, but the input and the output layers were modified according to the dimensionality of the new dataset. Such network has the disadvantage of having less training samples and more parameters, but potentially the advantage of integrating more information about the global structure of the system (e.g. distinguishing explicitly between different DOFs). The results are shown in Table \ref{tab:monolithic}.
\small
\begin{table}[h]
	\centering
	\begin{adjustbox}{max width=1\columnwidth,center}
		\begin{tabular}{|c||c|c|c|}
			\hline
			\textbf{Dataset} & \textbf{total RMSE} & \textbf{fit RMSE} & \textbf{Accuracy} \\
			\hline
			\hline
			\textbf{Training} & 4.0731 & 3.2723  & 60.20\% \\
			\hline
			\textbf{Validation} & 4.1597 & 3.3297 & 53.50\% \\
			\hline
			\textbf{Test }& 4.0531 & 3.2586 & 51.51\% \\
			\hline
		\end{tabular}
		\end{adjustbox}
	\caption{Training Results with a monolithic model}
	\label{tab:monolithic}
\end{table}
\normalsize
The performance is overall worse than in the modular case, especially for the classification accuracy, suggesting that a modular approach is advantageous with this problem.
\section{\uppercase{Conclusions and Future Work}}
\label{sec:conclusion}
In this work a method for posture control parameter identification based on CNN is presented. The system provides an efficient way to fit a model of the non-linear bio-inspired control system DEC on experimental data. This represents an advantage with respect to previous solutions relying on iterative methods. 
the training set is produced with parameters from normal distributions (although only the parameters producing a stable simulation are selected).  Future work will focus on the distribution of human data. An \textit{a posteriori} test can be performed comparing the distribution of the parameters identified on the test set with the expected distribution on real data. This can help the process of choosing between different possible network hyperparameters sets as shown in \cite{sforza2011rejection,sforza2013support}.

Some parameters were better identified than others as shown in Fig. \ref{fig:ErrorsCrop}. This may be due to the kind of experimental set up or due to the choice of the input features. For example the effect of $K_i$ is mainly visible just in the low pass components of the spectrogram. Furthermore, simulations and robot experiments are able to reproduce human behavior in the considered set up (PRTS support surface tilt) without the integral component of the neural controller \cite{Mergner2010,hettich2015human,Hettich2014}. The integral component of the controller is more important in analyzing transient behavior, e.g. reaching a desired position, where it can be used to guarantee zero tracking error \cite{ott2016good}. A tracking task may be better to identify $K_i$ more precisely.

The proposed CNN for posture control modeling can find application in  bio-inspired humanoid control, e.g. \cite{icinco07,icinco12,zebenay2015human,10.3389/fnbot.2018.00021}. The CNN can also be helpful in setting up wearable robots using the control parameters identified on the user \cite{icincoChugo.2019,Mergner2019}. The parameters can be a tool to benchmark humanoids and wearable devices \cite{torricelli2020benchmarking}, in particular, in the framework of the COMTEST project \cite{Lippi2019,lippi2020performance} that aims to make a posture control testbed available for the humanoid robotics community, and to define performance metrics.

\section*{\uppercase{Acknowledgements}}
\setlength{\intextsep}{-1pt}%
\setlength{\columnsep}{15pt}%
\begin{wrapfigure}{l}{0.08\columnwidth}
		{\includegraphics[width=0.12\columnwidth]{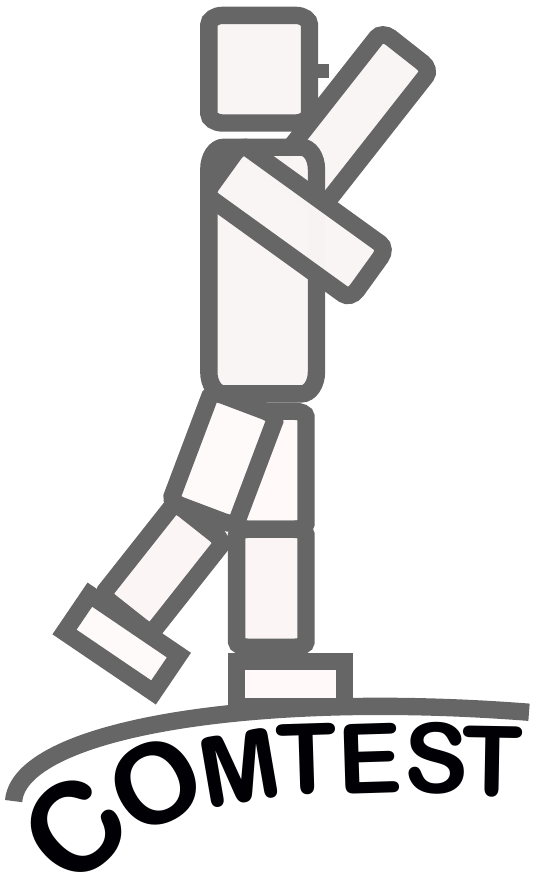}}
	\label{LOGO}
\end{wrapfigure}
\noindent This work is supported by the project COMTEST, a sub-project of EUROBENCH (European Robotic Framework for Bipedal Locomotion Benchmarking, www.eurobench2020.eu) funded by H2020 Topic ICT 27-2017 under grant agreement number 779963.

\bibliographystyle{apalike}
{\small
\bibliography{example}}

%

\end{document}